%% file: paper.tex
\documentclass{article}

\usepackage[preprint]{icml2026}

\input{assets/tex/preamble.tex}

\icmltitlerunning{Latent-Variable Learning of SPDEs via Wiener Chaos}
\begin{document}

\twocolumn[
  \icmltitle{Latent-Variable Learning of SPDEs via Wiener Chaos}



  \icmlsetsymbol{equal}{*}

  \begin{icmlauthorlist}
    \icmlauthor{Sebastian Zeng}{x}
    \icmlauthor{Andreas Petersson}{x}
    \icmlauthor{Wolfgang Bock}{x}
  \end{icmlauthorlist}

  \icmlaffiliation{x}{Department of mathematics and physics, Linnaeus University, Växjö, Sweden}

  \icmlcorrespondingauthor{Sebastian Zeng}{sebastian.zeng@lnu.se}

  \icmlkeywords{Stochastic Partial Differential Equations, Bayesian Latent Variable Models, Wiener Chaos Expansion}

  \vskip 0.3in
  ]



\printAffiliationsAndNotice{}  

\begin{abstract}
We study the problem of learning the law of linear stochastic partial differential equations (SPDEs) with additive Gaussian forcing from spatiotemporal observations. Most existing deep learning approaches either assume access to the driving noise or initial condition, or rely on deterministic surrogate models that fail to capture intrinsic stochasticity. We propose a structured latent-variable formulation that requires only observations of solution realizations and learns the underlying randomly forced dynamics. Our approach combines a spectral Galerkin projection with a truncated Wiener chaos expansion, yielding a principled separation between deterministic evolution and stochastic forcing. This reduces the infinite-dimensional SPDE to a finite system of parametrized ordinary differential equations governing latent temporal dynamics. The latent dynamics and stochastic forcing are jointly inferred through variational learning, allowing recovery of stochastic structure without explicit observation or simulation of noise during training. Empirical evaluation on synthetic data demonstrates state-of-the-art performance under comparable modeling assumptions across bounded and unbounded one-dimensional spatial domains.
\end{abstract}

\section{Introduction}
\input{sections/introductions}

\section{Method}
\input{sections/method}

\section{Experiments}
\input{sections/experiments}

\section{Discussion}
\input{sections/discussion}

\bibliography{paper}
\bibliographystyle{icml2026}

\newpage
\appendix
\onecolumn


\input{sections/suppmat.tex}

\end{document}

%% file: assets/tex/preamble.tex
\usepackage{wrapfig}
\usepackage{amssymb}
\usepackage{amsmath}
\usepackage{amsthm}
\usepackage{mathtools}
\usepackage{graphicx}
\usepackage{multirow}
\usepackage{float}         
\usepackage{adjustbox}
\usepackage{hyperref}
\usepackage[capitalise, noabbrev]{cleveref}
\usepackage{url}
\usepackage{subcaption}
\usepackage{microtype}
\usepackage{booktabs} 
\usepackage[disable]{todonotes}

\input{assets/tex/math_commands}

\input{assets/tex/colors}

\theoremstyle{plain}
\newtheorem{theorem}{Theorem}[section]
\newtheorem{proposition}[theorem]{Proposition}

\theoremstyle{definition}

\theoremstyle{remark}

%% file: assets/tex/math_commands.tex
\usepackage{amsmath,amsfonts,bm}

\def\eqref#1{equation~\ref{#1}}

\def\1{\bm{1}}

\DeclareMathAlphabet{\mathsfit}{\encodingdefault}{\sfdefault}{m}{sl}
\SetMathAlphabet{\mathsfit}{bold}{\encodingdefault}{\sfdefault}{bx}{n}



%% file: assets/tex/colors.tex
\definecolor{gred}{HTML}{DB4437}
\definecolor{gblue}{HTML}{4285F4}
\definecolor{ggreen}{HTML}{0F9D58}
\definecolor{gyellow}{HTML}{F4B400}

\definecolor{unired}{HTML}{A6192E}
\definecolor{uniblue}{HTML}{002D72}
\definecolor{unigreen}{HTML}{115740}

\definecolor{tabblue}{HTML}{1f77b4}
\definecolor{taborange}{HTML}{ff7f0e}
\definecolor{tabgreen}{HTML}{2ca02c}
\definecolor{tabred}{HTML}{d62728}
\definecolor{tabpurple}{HTML}{9467bd}
\definecolor{tabbrown}{HTML}{8c564b}
\definecolor{tabpink}{HTML}{e377c2}
\definecolor{tabgray}{HTML}{7f7f7f}
\definecolor{tabolive}{HTML}{bcbd22}
\definecolor{tabcyan}{HTML}{17becf}
\definecolor{tabpine}{HTML}{246C60}

\hypersetup{
    colorlinks=true,
    linkcolor=gblue,            
    citecolor=gblue,            
    filecolor=blue,             
    urlcolor=ggreen             
}

%% file: sections/introductions.tex
Stochastic partial differential equations (SPDEs) are fundamental for modeling complex systems under uncertainty. They capture the interplay of spatial interactions, temporal evolution, and random forcing in phenomena such as turbulent transport, fluctuating interfaces, and stochastic reaction-diffusion systems \citep{daPratoZabczyk92,Holden10,lord14}. In many applications, the governing equations and driving noise are only partially known, while data consist of spatiotemporal observations of solution realizations. This motivates the inverse problem of learning the underlying stochastic dynamics from data.

Learning the law of an SPDE, rather than individual trajectories, poses distinct challenges. The driving noise is typically infinite-dimensional, unobserved, and strongly coupled to the deterministic dynamics. A successful approach must therefore infer latent stochastic structure while respecting analytic properties such as linearity, spectral structure, and stability, which contrasts with much of the recent operator-learning literature.

Recent data-driven methods for PDEs include deep surrogate models and operator learning. Neural operators \citep{Kovachki23} learn mappings between infinite-dimensional function spaces and perform strongly for deterministic solution operators, e.g., Fourier Neural Operators \citep{Li20a} and DeepONets \citep{Lu21a}. However, standard neural operators are deterministic, and extending them to SPDE dynamics is challenging since randomness must be modeled explicitly.

Many stochastic extensions learn noise-conditioned solution operators and thus assume access to noise realizations or finite-dimensional noise features. This includes Neural SPDEs \citep{Salvi22}, inspired by neural controlled differential equations \citep{Kidger20}, as well as neural SPDE solvers for uncertainty quantification \citep{Beauchamp23} and the Deep Latent Regularity Network \citep{Gong23}. These methods learn conditional solution maps given noise features, rather than inferring the law when the noise is latent.

A related line of work combines operator learning with chaos-based representations. \citet{Shi26} combine Wiener chaos expansions with neural operators for noise-conditioned learning on bounded domains, while \citet{Neufeld24} parameterize deterministic chaos coefficients using neural networks assuming known Gaussian chaos variables. \citet{Petursson25} propose a chaos-inspired framework that focuses on the expected values of SPDE solutions. Such moment-based targets are useful for uncertainty quantification, but they do not identify the full pathwise law and may miss intrinsic variability central to the inverse setting.

In this work, we propose a structured latent-variable framework for learning linear SPDEs with additive Gaussian forcing from spatiotemporal observations alone. Our approach uses spectral Galerkin projections and Wiener chaos expansions to construct a finite-dimensional representation that separates deterministic evolution from stochastic excitation in a principled manner. This representation reduces the infinite-dimensional SPDE to a system of parametrized ordinary differential equations governing latent temporal modes, while retaining explicit control over stochastic forcing. Crucially, the latent dynamics and stochastic components are inferred jointly through variational learning, without requiring access to noise realizations or explicit simulation of the driving process during training. This enables recovery of the stochastic law underlying the observed dynamics rather than simply fitting individual trajectories. Through numerical experiments on synthetic data, we demonstrate that the proposed method achieves state-of-the-art performance under comparable modeling assumptions across both bounded and unbounded one-dimensional spatial domains.

%% file: sections/method.tex
\subsection{Notation and problem setting}
\label{subsec:problem}

We introduce the method using a deliberately simple and well-understood
stochastic partial differential equation.
Our goal is \emph{not} to propose new analytical solution techniques for this
equation, but to use it as a transparent testbed that allows us to clearly
explain and validate the proposed learning framework.

Let $H$ be a separable Hilbert space and consider the linear SPDE
\begin{equation}
\label{eq:spde}
\begin{cases}
dX_t = A X_t\,dt + dW_t, \quad t\in[0,T],\\
X_0 = u_0 \in H,
\end{cases}
\end{equation}
where $A : \mathcal D(A)\subset H \to H$ is a linear operator generating a bounded (up to time $T$) strongly continuous semigroup $(e^{tA})_{t\ge 0}$, $(W_t)_{t\ge 0}$ is a $Q$-Wiener process and $u_0$ is deterministic. The covariance operator $Q$ is self-adjoint and nonnegative. 
We further assume that $Q$ is trace-class to ensure that a solution exists regardless of any additional properties of $A$.

Under these assumptions, the equation \labelcref{eq:spde} admits the mild solution
\begin{equation}
\label{eq:mild}
X_t = e^{tA}u_0 + \int_0^t e^{(t-s)A}\,dW_s,
\end{equation}
which we use solely as a conceptual reference.
Throughout this work, the equation serves as a controlled setting in which
the interaction between stochasticity, model reduction, and learning can be
studied in isolation.

For simplicity and clarity, we further assume that $-A$ is diagonalizable in
an orthonormal basis $\{h_n\}_{n\ge1}\subset H$ with
\begin{equation}
\label{eq:eigen}
-Ah_n = \lambda_n h_n, \qquad 0<\lambda_1\le\lambda_2\le\cdots.
\end{equation}
This assumption is standard and allows for an explicit and intuitive
finite-dimensional representation of the dynamics.

The stochastic analytic background used throughout this section is classical.
Existence and uniqueness of mild solutions to linear SPDEs with additive
Gaussian noise, as well as spectral Galerkin approximations, are well covered
in standard monographs such as \citet{daPratoZabczyk92,lord14}.
The Wiener--It\^o chaos expansion and the associated propagator formulation
for linear stochastic evolution equations follow the classical framework of
\citet{Lototsky06}.

We include the present derivation solely for completeness and pedagogical
clarity, as it provides a transparent link between stochastic modeling and
deterministic latent dynamics.

\subsection{Galerkin reduction in space}
\label{subsec:galerkin}

As a first reduction step, we discretize the spatial component of the SPDE
using a spectral Galerkin projection.
For $N\ge1$, define the finite-dimensional subspace
\[
V_N := \mathrm{span}\{h_1,\dots,h_N\}
\]
and let $P_N : H \to V_N$ denote the orthogonal projection.

Projecting the solution process onto $V_N$ yields
$\widetilde X_t := P_N X_t$, which satisfies the finite-dimensional stochastic
evolution equation
\begin{equation}
\label{eq:galerkin}
d\widetilde X_t = A_N \widetilde X_t\,dt + d\widetilde W_t,
\end{equation}
where $A_N := P_N A P_N$ and $\widetilde W_t := P_N W_t$ is a $Q_N$-Wiener process
with covariance $Q_N=P_NQP_N$.
Since $A_N$ is a bounded operator on $V_N$, the equation admits the explicit
representation
\begin{equation}
\label{eq:galerkin_mild}
\widetilde X_t = e^{tA_N}\widetilde X_0 + \int_0^t e^{(t-s)A_N}\,d\widetilde W_s.
\end{equation}

This step transforms the infinite-dimensional SPDE into a finite-dimensional
stochastic system while preserving its essential structure.

\subsection{Wiener--It\^o chaos expansion}
\label{subsec:chaos}

To make the stochastic dependence explicit, we expand $\widetilde X_t$ using
a Wiener--It\^o chaos decomposition.
Let $\{m_k\}_{k\ge1}$ be an orthonormal basis of $L^2([0,T])$ and let
$\{w_t^\ell\}_{\ell\ge1}$ be independent standard Wiener processes driving $W$.
Define the scalar Gaussian random variables
\begin{equation}
\label{eq:xi_kl}
\xi_k^\ell := \int_0^T m_k(s)\,dw_s^\ell .
\end{equation}

Using these variables, we construct an orthonormal basis of
$L^2(\Omega,\mathcal F_T,\mathbb P)$ indexed by multi-indices
$\alpha=(\alpha_k^\ell)$ with finite order $|\alpha|$, given by
\begin{equation}
\label{eq:xi_alpha}
\xi_\alpha := \frac{1}{\sqrt{\alpha!}}
\prod_{k,\ell} H_{\alpha_k^\ell}(\xi_k^\ell),
\end{equation}
where $H_n$ denotes the probabilists' Hermite polynomial. Note that $\xi_0 = 1$ while $(\xi_\alpha : |\alpha|=1)$ is a set of independent $\mathcal{N}(0,1)$ random variables. This follows from $H_0(x) = 1$ and $H_1(x) = x$ for $x \in \mathbb{R}$.

Consequently, for each fixed $t$, the $V_N$-valued random variable
$\widetilde X_t$ admits the expansion
\begin{equation}
\label{eq:chaos_expansion}
\widetilde X_t(\omega)
=
\sum_{\alpha} \widetilde X_t^{(\alpha)}\,\xi_\alpha(\omega),
\qquad
\widetilde X_t^{(\alpha)} := \mathbb E[\widetilde X_t\,\xi_\alpha]\in V_N.
\end{equation}

This representation separates randomness, encoded in the fixed basis functions
$\xi_\alpha(\omega)$, from time-dependent deterministic coefficients.

\subsection{Propagator equations}
\label{subsec:propagator}

The coefficient functions
$
\{\widetilde X_t^{(\alpha)}\}_{\alpha}
$
are referred to as the \emph{propagators}.
They describe how each stochastic mode evolves over time and form the core
objects of our learning approach.

Multiplying \labelcref{eq:galerkin} by $\xi_\alpha$ and taking expectations yields
\begin{equation}
\label{eq:prop_general}
\frac{d}{dt}\widetilde X_t^{(\alpha)}
= A_N \widetilde X_t^{(\alpha)}
+ \frac{d}{dt}\mathbb E[\widetilde W_t\,\xi_\alpha].
\end{equation}
Due to orthogonality of the Wiener--It\^o chaos basis, stochastic forcing enters
only through first-order chaos terms.
As a result, the system of propagators closes at first order.

\begin{proposition}[First-order closure]
\label{prop:closure}
For linear SPDEs with additive Gaussian noise, the propagator coefficients
$\widetilde X_t^{(\alpha)}$ vanish identically for all $|\alpha|\ge2$.
\end{proposition}

Writing the propagators in the eigenbasis of $A$,
$\widetilde X_t^{(\alpha,n)} := \langle \widetilde X_t^{(\alpha)}, h_n\rangle_H$,
leads to a system of explicit ordinary differential equations:
\begin{equation}
\label{eq:prop_mean}
\begin{cases}
\tfrac{d}{dt}\widetilde X_t^{(0,n)} = -\lambda_n \widetilde X_t^{(0,n)}, \quad t\in[0,T],\\[4pt]
 \widetilde X_0^{(0,n)} = \langle u_0, h_n\rangle_H,
\end{cases}
\end{equation}
and
\begin{equation}
\label{eq:prop_first}
\begin{cases}
\begin{aligned}
\tfrac{d}{dt}\widetilde X_t^{(\alpha,n)} =
&-\lambda_n \widetilde X_t^{(\alpha,n)}
+ \mathbf 1_{\{\alpha=e_{k^\ast,\ell^\ast}\}} m_{k^\ast}(t) \\
&\quad\times \langle Q^{1/2}h_{\ell^\ast},h_n\rangle_H,
\quad t\in[0,T],
\end{aligned}\\[4pt]
\widetilde X_0^{(\alpha,n)} = 0,
\end{cases}
\end{equation}
for $n=1,\dots,N$, where for a pair $(k,\ell)$, we use the notation
\[
(e_{k,\ell})_{k',\ell'} =
\begin{cases}
1, & (k',\ell') = (k,\ell),\\
0, & \text{otherwise}.
\end{cases}.
\]

\subsection{Truncation and reconstruction}
\label{subsec:reconstruction}

Restricting to first-order indices
$\mathcal J_{K,L} := \{e_{k,\ell} : 1\le k\le K,\,1\le \ell\le L\}$,
the latent dimension of the reduced system is
\[
d = N(1+KL).
\]
The corresponding truncated approximation of the solution process is given by
\begin{equation}
\label{eq:reconstruct}
\begin{aligned}
X_{N,K,L}(t,\omega)
&=
\sum_{n=1}^N \widetilde X_t^{(0,n)} h_n
\\
&\quad+
\sum_{n=1}^N\sum_{\alpha\in\mathcal J_{K,L}}
\widetilde X_t^{(\alpha,n)} h_n\,\xi_\alpha(\omega).
\end{aligned}
\end{equation}

This form makes explicit how the observed stochastic solution is generated from
a finite set of deterministic propagator trajectories.
The clear separation between stochastic features and temporal dynamics is what
enables a latent-variable learning formulation, which we describe next.

\vspace*{-.3cm}
 \paragraph{Remark on Truncation and Approximation.}
The truncation levels $N$ (spatial modes) and $(K,L)$ (first-order chaos indices) yield latent dimension $d = N(1+KL)$.
Classical Galerkin approximation results and Wiener chaos truncation bounds imply that the approximation error decreases as $N\to\infty$ and $K,L\to\infty$ under standard regularity assumptions; see, e.g., \citet{lord14,Lototsky06}.
Thus, learning the finite propagator dynamics is a controlled and theoretically justified surrogate for the original SPDE.

\subsection{A latent-variable model}

In our approach, learning the solution process $X_t$ of the linear SPDE \labelcref{eq:spde} from discretized sample paths is formulated as a joint inference problem over two latent components: the temporal evolution of the propagator coefficients $\{\widetilde{X}_t^{\alpha}\}_{\alpha}$ and the associated chaos coordinates $\{\xi_{\alpha}(\omega)\}_{\alpha}$, both inferred from discretized i.i.d.\ realizations $X_t(\omega)$. Although neither component is directly observable, their structure is partially known \emph{a priori}: by \labelcref{eq:prop_mean,eq:prop_first}, the propagators evolve according to a deterministic evolution equation, and the first-order chaos coordinates $\xi_{\alpha}$ where $|\alpha|=1$ are independent standard Gaussian random variables. We therefore model both quantities as latent variables and infer them within a variational Bayesian framework, enabling parameter estimation from partially observed sample paths.

We adopt the latent ODE framework of \citet{rubanova19} to construct our model.
Let $\{\mathbf{X}^{m_1}\}_{m_1=1}^{M_1}$ denote i.i.d.\ realizations of the solution process,
each partially observed on a discrete spatiotemporal mesh
\[
\{(t_{m_2}, x_{m_3})\}_{m_2=0,\,m_3=1}^{M_2,\,M_3}
\subset [0,T] \times \mathcal{D},
\qquad \mathcal{D} \subset \mathbb{R}^D.
\]
An encoder network $\texttt{Enc}_{\boldsymbol{\phi}}$ infers a joint approximate posterior

\[ q_{\boldsymbol{\phi}}(\boldsymbol{z}_{t_0}, \boldsymbol{\xi} \mid \mathbf{X}^{m_1}),
\]
over the initial latent state $\boldsymbol{z}_{t_0}$ and the latent chaos coordinate
vector $\boldsymbol{\xi} = (\xi_{\alpha})_{\alpha\in\mathcal{J}_{K,L}}$.
Here, $\boldsymbol{z}_{t_0}$ parameterizes the initial values of the zero-order propagator coefficients.
Although the underlying SPDE has a deterministic initial condition shared across realizations, we keep $\boldsymbol{z}_{t_0}$ in the variational family to accommodate partial observations and measurement noise; in the deterministic limit the posterior concentrates and recovers a shared initial state.

Given the latent initial condition $\boldsymbol{z}_{t_0}$, the temporal evolution
of the propagator coefficients is governed by the system of ordinary differential
equations \labelcref{eq:prop_mean,eq:prop_first}.
The dynamics are parameterized by $\boldsymbol{\theta} = (\boldsymbol{\lambda},
\boldsymbol{q})$, where $-\boldsymbol{\lambda}$ parameterizes the deterministic drift
and $\boldsymbol{q}$ encodes the stochastic forcing.

These parameters enter through a vector field $f_{\boldsymbol{\theta}}$, which is
integrated numerically on the time grid $\{t_{m_2}\}_{m_2=1}^{M_2}$.
For $m_2 = 1,\dots,M_2$, we write the resulting latent states as
\[
\boldsymbol{z}_{t_{m_2}}
=
\mathrm{ODESolve}\!\left(
f_{\boldsymbol{\theta}},\, \boldsymbol{z}_{t_0},\, t_{m_2}
\right),
\]
 while $\boldsymbol{z}_{t_0}$ is produced directly by the encoder (without solving the ODE), and $\boldsymbol{\theta}$ governs the subsequent latent evolution through $f_{\boldsymbol{\theta}}$.

Since the latent dynamics evolve in a reduced space, we introduce a reconstruction
operator
\[
\mathcal{R}:
\bigl(\{\boldsymbol{z}_{t_{m_2}}\}_{m_2=0}^{M_2},\, \boldsymbol{\xi}\bigr)
\;\mapsto\;
\widehat{\mathbf{X}} \in \mathbb{R}^{(M_2+1) \times M_3},
\]
that maps the latent trajectory and chaos coordinates to the observation space.

For convenience, we denote by $\mathcal{R}_{m_2}(\boldsymbol{z}_{t_{m_2}},\boldsymbol{\xi}) := \widehat{\mathbf{X}}_{m_2,:}$ the time-slice reconstruction at $t_{m_2}$.

This defines a likelihood model
\[
p_{\boldsymbol{\gamma}}\!\left(
\mathbf{X}^{m_1}
\,\middle|\,
\mathcal{R}\bigl(\{\boldsymbol{z}_{t_{m_2}}\}_{m_2=0}^{M_2}, \boldsymbol{\xi}\bigr)
\right),
\]
parameterized by a decoder network $\texttt{Dec}_{\boldsymbol{\gamma}}$, and enables
end-to-end training within the variational framework.

Given a prior distribution \(p(\boldsymbol{z}_{t_0}, \boldsymbol{\xi})\), we define the evidence lower bound (ELBO) as

\vspace{-0.2cm}
\begin{equation}
\label{eq:elbo}
\begin{array}{@{}l@{}}
\mathcal{L}(\boldsymbol{\phi},\boldsymbol{\theta}, \boldsymbol{\gamma}; \mathbf{X}^{m_1})=
\\[0.2em]
\multicolumn{1}{@{}r@{}}{\displaystyle
\mathbb{E}_{q_{\boldsymbol{\phi}}(\boldsymbol{z}_{t_0}, \boldsymbol{\xi} \mid \mathbf{X}^{m_1})}
\!\left[
\sum_{m_2=0}^{M_2}
\log
p_{\boldsymbol{\gamma}}\!\left(
\mathbf{X}^{m_1}_{m_2,:}
\,\middle|\,
\mathcal{R}_{ m_2}(\boldsymbol{z}_{t_{m_2}}, \boldsymbol{\xi})
\right)
\right]
}
\\
\noalign{\vskip 0.5em} 
\multicolumn{1}{@{}r@{}}{\displaystyle
-\mathrm{KL}\!\left(
q_{\boldsymbol{\phi}}(\boldsymbol{z}_{t_0}, \boldsymbol{\xi} \mid \mathbf{X}^{m_1})
\Vert
p(\boldsymbol{z}_{t_0}, \boldsymbol{\xi})
\right)
}
\end{array}
\end{equation}
where $\mathbf{X}^{m_1}_{m_2,:} \in \mathbb{R}^{M_3}$ denotes the observed spatial field at time $t_{m_2}$. 

Model training is performed by maximizing the ELBO with respect to the
variational and generative parameters, i.e.,
\begin{equation}
\label{eq:elbo_opt}
(\boldsymbol{\phi}, \boldsymbol{\theta}, \boldsymbol{\gamma})
=
\arg\max_{\boldsymbol{\phi}, \boldsymbol{\theta}, \boldsymbol{\gamma}}
\;
\mathcal{L}(\boldsymbol{\phi}, \boldsymbol{\theta}, \boldsymbol{\gamma}; \mathbf{X}^{m_1}).
\end{equation}
At the dataset level, this objective is aggregated over
\(\{\mathbf{X}^{m_1}\}_{m_1=1}^{M_1}\)
and optimized using stochastic gradient-based methods with minibatches.

\subsection{Latent-variable structure and factorization}
We assume a factorized prior
\[
p(\boldsymbol{z}_{t_0}, \boldsymbol{\xi})
=
p(\boldsymbol{z}_{t_0})\,p(\boldsymbol{\xi}),
\]
and employ a mean-field variational approximation
\[
q_{\boldsymbol{\phi}}(\boldsymbol{z}_{t_0}, \boldsymbol{\xi}\mid \mathbf{X}^{m_1})
=
q_{\boldsymbol{\phi}}(\boldsymbol{z}_{t_0}\mid \mathbf{X}^{m_1})\,
q_{\boldsymbol{\phi}}(\boldsymbol{\xi}\mid \mathbf{X}^{m_1}).
\]
This choice is standard in variational inference and enables scalable
optimization, at the expense of neglecting posterior dependencies between the
initial propagator coefficients and the stochastic forcing coordinates.

We emphasize that in our problem setting, the latent variables
$\boldsymbol{z}_{t_0}$ and $\boldsymbol{\xi}$ differ fundamentally in their
dependence on the observed data.
The initial latent state $\boldsymbol{z}_{t_0}$ parameterizes the deterministic
evolution of the propagator coefficients and is therefore shared across all
observed solution trajectories.
{In our implementation, $q_{\phi}(\boldsymbol{z}_{t_0}\mid \mathbf{X}^{m_1})$ is computed per realization, but the learned posteriors are encouraged by the data and KL regularization to concentrate around a common value across $m_1$.
}
In contrast, the chaos coordinate vector $\boldsymbol{\xi}$ encodes the
stochastic forcing and is realization-specific, varying across sample paths
$m_1$.

Despite this distinction, both latent variables are modeled probabilistically.
While $\boldsymbol{z}_{t_0}$ corresponds to a deterministic initial condition in
the underlying SPDE \labelcref{eq:spde}, the variational formulation allows the posterior to
concentrate as data are observed, effectively recovering a degenerate
distribution in the deterministic limit.
At the same time, this formulation naturally accommodates measurement noise in
the observations and provides a principled extension to settings with randomized
or uncertain initial conditions.

\subsection{Priors and variational family}
We adopt standard Gaussian priors \(\mathcal{N}(\mathbf{0}, I)\) for all latent variables;
for the chaos variable \(\boldsymbol{\xi}\), this choice is justified by Wiener chaos
theory, since the first-order chaos basis functions \((\xi_\alpha : |\alpha| = 1)\)
associated with a \(Q\)-Wiener process are independent standard Gaussian random variables.

The approximate posterior distributions
\(q_{\boldsymbol{\phi}}(\boldsymbol{z}_{t_0}\mid \mathbf{X}^{m_1})\) and
\(q_{\boldsymbol{\phi}}(\boldsymbol{\xi}\mid \mathbf{X}^{m_1})\) are modeled as multivariate Gaussian distributions with diagonal covariance.
Specifically, an encoder network \(\texttt{Enc}_{\boldsymbol{\phi}}\) maps each observed
sample path \(\mathbf{X}^{m_1}\) to the variational parameters
\[
(\boldsymbol{\mu}_z, \boldsymbol{\sigma}^{2}_z)
\in \mathbb{R}^{d_z} \times \mathbb{R}^{d_z},
\qquad
(\boldsymbol{\mu}_{\xi}, \boldsymbol{\sigma}^{2}_{\xi})
\in \mathbb{R}^{d_{\xi}} \times \mathbb{R}^{d_{\xi}},
\]
respectively.

\subsection{Reconstruction operator and observation model}
We next describe how the inferred latent variables are mapped back to the
observation space.
The reconstruction operator $\mathcal{R}$ implements the evaluation of the truncated approximation of the solution process \labelcref{eq:reconstruct} on a discretized observation mesh.
Given fixed eigenfunctions $\{h_n\}_{n=1}^{N}$ and spatial mesh points
$\{x_{m_3}\}_{m_3 = 1}^{M_3}$, $\mathcal{R}$ maps the latent trajectory
$\{\boldsymbol{z}_{t_{m_2}}\}_{m_2=0}^{M_2}$ and the chaos coordinate vector
$\boldsymbol{\xi} = (\xi_{\alpha})_{\alpha\in\mathcal{J}_{K,L}}$
 to discretized sample paths in the
observation space.

Specifically, for each time point $t_{m_2}$ and spatial location $x_{m_3}$,
the reconstructed field is obtained by evaluating the truncated approximation,
\[
\begin{aligned}
\widehat{X}_{m_2,m_3}
&=
\sum_{n=1}^N z^{(0)}_{n}(t_{m_2})\, h_n(x_{m_3})
\\
&\quad+\;
\sum_{n=1}^N \sum_{\alpha\in\mathcal J_{K,L}}
z^{(\alpha)}_{n}(t_{m_2})\, h_n(x_{m_3})\, \xi_{\alpha}.
\end{aligned}
\]
where the latent coefficients
$\{z^{(\alpha)}_{n}(t)\}$ are components of the latent state
$\boldsymbol{z}_t$ associated with chaos index $\alpha$ and spatial mode $n$.

This reconstruction bridges the reduced latent dynamics and the observation
space while preserving the structure imposed by the Wiener chaos expansion
and the Galerkin projection.

The decoder $\texttt{Dec}_{\boldsymbol{\gamma}}$ parameterizes a Gaussian observation model,
\[
p_{\boldsymbol{\gamma}}(\mathbf{X}^{m_1}_{m_2,:} \mid \widehat{\mathbf{X}}_{m_2,:})
=
\mathcal{N}\!\left(
\widehat{\mathbf{X}}_{m_2,:},
\operatorname{diag}\bigl(\boldsymbol{\sigma}^2\bigr)
\right),
\]
where $\boldsymbol{\sigma}^2 \in \mathbb{R}^{M_3}$ denotes a vector of learnable variance
parameters, one per spatial observation location.
The variance is shared across time, reflecting time-homogeneous observation noise, while
allowing for spatially varying uncertainty.
This structured likelihood reduces the number of free parameters and improves stability
in the variational optimization.

\subsection{Encoder architecture}

The encoder network $\texttt{Enc}_{\boldsymbol{\phi}}$ amortizes inference over
the latent initial state $\boldsymbol{z}_{t_0}$ and the chaos coordinates
$\boldsymbol{\xi}$ from discretized spatiotemporal observations.
In contrast to \citet{rubanova19}, who employ a neural ODE to parameterize the
encoder, we adopt an attention-based encoder architecture based on the
multi-time attention network (mTAN) \citep{Shukla21a}, which is well suited for
irregularly sampled time series.

In the present implementation, we focus on one-dimensional spatial domains,
such that each observation $\mathbf{X}^{m_1}_{m_2,:}$ is represented as a vector
in $\mathbb{R}^{M_3}$.
This choice is dictated by the encoder architecture and does not reflect a
conceptual limitation of the latent-variable formulation.
Extending the approach to higher-dimensional spatial domains would require
replacing the encoder with a suitable architecture capable of handling
higher-dimensional spatial structure, without modifying the latent dynamics
or reconstruction operator.

%% file: sections/experiments.tex
\subsection{Experimental setup and regimes}
In this section, we evaluate the proposed model on linear SPDEs posed on both bounded and unbounded spatial domains.
Across all experiments, we observe solution trajectories on a discrete spatiotemporal mesh.
In contrast to many existing approaches, the realization of the driving noise is never observed and is treated as a latent variable to be inferred from solution data alone.
This setting allows us to assess whether the stochastic structure of the underlying SPDE can be recovered from solution observations alone.

We consider two experimental regimes that differ in domain geometry, boundary conditions, and the associated spectral basis, while sharing identical observation assumptions.
An overview of the regimes is given in \cref{tbl:exp:regimes}.

\begin{table*}[t]
  \caption{Overview of the two experimental regimes, highlighting differences in domain geometry, operator structure, and spectral representation, while maintaining identical observation assumptions.}
  \label{tbl:exp:regimes}
  \begin{center}
    \begin{small}
      \begin{sc}
        \begin{tabular}{lcc}
          \toprule
                            & Regime A         & Regime B   \\
          \midrule
          SPDE              
          & Ornstein--Uhlenbeck 
          & Stochastic heat equation
          \\
          State space $\mathcal{H}$     
          & $L^2(\mathbb{R}, \gamma)$ 
          & $L^2(0,\pi)$
          \\
          Boundary conditions    
          & -- 
          & Dirichlet
          \\
          Generator $A$ 
          & $(Au)(x) = u''(x) - x u'(x) - u(x)$ 
          & $(Au)(x) = u''(x)$
          \\
          Spectral basis $\{h_n\}_{n \geq 1}$ 
          & Probabilists' Hermite polynomials 
          & Fourier sine functions
          \\
          Driving noise $(W_t)_{t \geq 0}$ 
          & \multicolumn{2}{c}{$Q$-Wiener process}
          \\
          Observations       
          & \multicolumn{2}{c}{$\{ X(t_{m}, \cdot) \}_{m=1}^{M_1}$ (solution trajectories only; no noise realizations)}
          \\
          \bottomrule
        \end{tabular}
      \end{sc}
    \end{small}
  \end{center}
  \vskip -0.1in
\end{table*}

\subsection{Common setup across regimes}

\paragraph{Solution representation.}
We approximate the mild solution \(X_t\) in \eqref{eq:mild} by the truncated representation \eqref{eq:reconstruct}.
We retain \(N=8\) spatial modes, which capture the dominant energy of \(X_t\) under the chosen noise spectrum.
For the stochastic forcing, we use \(L=8\) noise components and \(K=16\) modes in the Wiener chaos truncation.
\vspace*{-.3cm}
\paragraph{Model parameterization.}
The truncated representation \eqref{eq:reconstruct} reduces the infinite-dimensional SPDE to a finite collection of time-dependent coefficient processes.
Our model provides a structured, data-driven parameterization of the coefficient processes \(\{\widetilde X_t^{(\alpha,n)}\}\) and the associated stochastic variables, enabling their joint inference from observed solution trajectories.
\vspace*{-.3cm}
\paragraph{Model and training.}
Across both regimes, all models are trained using the same optimization protocol.
We employ the Adam optimizer \citet{Kingma15} with separate base learning rates for the encoder, latent dynamics, and decoder, using a linear warmup over the first \(10\) epochs followed by cosine learning-rate scheduling.
The latent dynamics are integrated using a fourth-order Runge--Kutta (RK4) method during both training and evaluation, implemented via the \texttt{torchdiffeq} library \citep{torchdiffeq}.
Models are trained for \(2000\) epochs, which is sufficient for convergence under the chosen learning-rate schedule.
Training data consist of \(1000\) solution trajectories, split into \(70\%\) training, \(15\%\) validation, and \(15\%\) test sets.
Model selection is performed by selecting the checkpoint achieving the best performance on the validation set.
We apply weight decay and KL regularization on the latent variables to stabilize training.
Concrete training and model hyperparameters are reported in Appendix~\labelcref{app:sec:hyperparameters}.
\vspace*{-.3cm}
\paragraph{Latent dynamics initialization.}
The parameters governing the latent dynamics are initialized to ensure stable evolution at the start of training.
All drift coefficients \(\lambda_n\) are initialized to \(1\).
For the stochastic forcing, we assume for simplicity that the covariance operator \(Q\) is diagonal in the same eigenbasis as the generator \(A\); the corresponding noise amplitudes \(q_n\) are likewise initialized to \(1\).
During training, both the eigenvalues $\lambda_n$ (entering the drift as $-\lambda_n$) and the noise coefficients $q_n$ are constrained to remain strictly positive to ensure stability of the latent dynamics.
This neutral and stable initialization was sufficient for reliable optimization across all experiments.
\vspace*{-.3cm}
\paragraph{Data generation.}
We first describe common assumptions used for data generation across both regimes; regime-specific generators, initial conditions, and domains are specified below.
To generate training and test data, we simulate realizations of the mild solution to a linear SPDE defined in \labelcref{subsec:problem}.
For simplicity, we assume that the covariance operator \(Q\) diagonalizes in the eigenbasis \(\{h_n\}_{n\geq 1}\) of \(A\), i.e., \(Q h_n = q_n h_n\).
We choose \(q_n = n^{-(2r + 1 + \varepsilon)}\) with \(r = 0.5\) and \(\varepsilon = 0.01\), ensuring that \(Q\) is trace-class.
Given \(M_1 = 1000\) independent samples, we generate solution trajectories by computing the truncated mild solution on a discrete spatiotemporal mesh
\[
\{(t_{m_2}, x_{m_3})\}_{m_2=0,\; m_3=1}^{M_2,\; M_3}
\subset [0,T] \times \mathcal{D}, \qquad \mathcal{D} \subset \mathbb{R}.
\]
The reference solutions are obtained using a spectral Galerkin discretization in space combined with a semi-implicit Euler--Maruyama scheme in time, as commonly used in the numerical approximation of linear SPDEs; see, e.g., \citet{lord14}.

\subsection{Evaluation metrics}

We evaluate model performance using both trajectory-level accuracy metrics and distributional criteria to assess recovery of the stochastic structure.
\vspace*{-.3cm}
\paragraph{Trajectory accuracy.}
Prediction accuracy is quantified using the relative \(L^2\) error and the root mean
squared error (RMSE), computed between predicted and reference solution trajectories
on the observation grid. Errors are averaged over time, space, and test samples,
excluding the initial time point \(t_0\).
\vspace*{-.3cm}
\paragraph{Distributional accuracy.}
To assess whether the learned model captures the law of the solution process, we evaluate second-order statistics induced by the learned dynamics.
In particular, we compare empirical variances of both the reconstructed solution fields and their modal coefficients, estimated from samples drawn from the learned model, to corresponding statistics computed from reference test trajectories.
Unless stated otherwise, all distributional evaluations are performed using unconditional samples drawn from the learned model.

\subsection{Regime A: Unbounded domain with Ornstein--Uhlenbeck dynamics}

Let \(H := L^2(\mathbb{R}, \gamma)\), where
\[
\gamma(x) = (2\pi)^{-1/2} e^{-x^2/2}
\]
denotes the standard Gaussian density, and let \(A\) denote the (perturbed)
Ornstein--Uhlenbeck operator
\[
(Au)(x) = u''(x) - x u'(x) - u(x).
\]
The operator \(-A\) admits a complete orthonormal eigenbasis
\((h_n)_{n\ge1}\) of \(H\), given by the normalized probabilists' Hermite
polynomials. Specifically,
\[
\scalebox{0.92}{$
h_n(x) = \frac{1}{\sqrt{n!}}\,\mathrm{He}_n(x),
\qquad
\mathrm{He}_n(x)
=
(-1)^n e^{x^2/2} \frac{d^n}{dx^n} e^{-x^2/2},
$}
\]
with corresponding eigenvalues
\[
-A h_n = \lambda_n h_n, \qquad \lambda_n = n+1 .
\]

We study the stochastic evolution equation
\begin{equation}
\label{exp:eq:ou}
\begin{cases}
dX_t = A X_t\,dt + dW_t, \quad t\in[0,T],\\
X_0 = u_0 \in H,
\end{cases}
\end{equation}
where \((W_t)_{t\ge0}\) is an \(H\)-valued \(Q\)-Wiener process.

We fix the final time \(T=1\) and choose a deterministic initial condition
\(u_0 \in H\) given by
\[
u_0(x) = 10\,\exp\!\left(-\frac{x^2}{2\sigma^2}\right),
\qquad \sigma = 0.8.
\]

The time interval \([0,T]\) is discretized using \(M_2=200\) points.
For evaluation on a discrete spatial mesh, we restrict to the interval
\([-3,3]\subset\mathbb{R}\) and use \(M_3=200\) uniformly spaced grid points.
This choice yields a spatial resolution comparable to the temporal resolution.
All solution trajectories are generated using the numerical procedure described
in the data generation setup.

\begin{table}[H]
 \caption{Prediction accuracy on Regime A (Ornstein--Uhlenbeck dynamics) measured by relative $L^2$ error and RMSE (lower is better). Reported values are mean $\pm$ standard deviation over test trajectories, averaged over 3 random model initializations.}
 \label{tbl:exp:regime_a:results}
 \begin{center}
   \begin{small}
     \begin{sc}
       \scalebox{0.96}{
       \begin{tabular}{lr|cc@{}}
         \toprule
         \textbf{Model} & \#Para & rel. $L^2 \;(\times 10^{-1})$ & RMSE $(\times 10^{-1})$ \\
         \midrule
         SV \& NN & 1408   & 2.276 $\pm$ 0.007 & 7.885 $\pm$ 0.009 \\
         \midrule
         Ours     & 600864 & \textbf{0.779} $\pm$ 0.031 & \textbf{2.615} $\pm$ 0.110 \\
         \bottomrule
       \end{tabular}
       }
     \end{sc}
   \end{small}
 \end{center}
 \vskip -0.2in
\end{table}

Table~\ref{tbl:exp:regime_a:results} reports prediction accuracy for Regime~A.
We compare our approach to the SV \& NN baseline of \citet{Neufeld24}, using the authors’ publicly available implementation \cite{neufeld24chaosspde_code}, adapted to our experimental setup with hyperparameters kept fixed.
Since this model assumes access to the driving noise (or equivalently the chaos coordinates), which is unavailable when only solution paths are observed, we sample the chaos coordinates independently from their prescribed marginal distribution, namely a standard Gaussian.
Under identical observation assumptions, our method substantially improves both relative $L^2$ error and RMSE.

To assess approximation of the solution law in Regime~A, we compare second-order
statistics of generated and reference trajectories. Specifically, we examine the
temporal evolution of the spatially averaged variance, as well as the modal energy
spectrum induced by projection onto the probabilists' Hermite basis.
Formal definitions of these law-level diagnostics are provided in
Appendix~\ref{sec:evaluation_metrics}.
As shown in \cref{fig:regimeA_law}, both the variance growth and the distribution of
energy across spatial modes are accurately reproduced by the learned model,
indicating that it captures the correct second-order stochastic structure of the
solution process.

Exact identification of individual spectral parameters is not expected, since the decomposition of uncertainty into process noise and reconstruction noise is not identifiable from partial observations. Nevertheless, the induced field law is faithfully matched.

\begin{figure*}[t]
  \centering

  \begin{subfigure}[t]{0.32\textwidth}
    \centering
    \includegraphics[width=\textwidth]{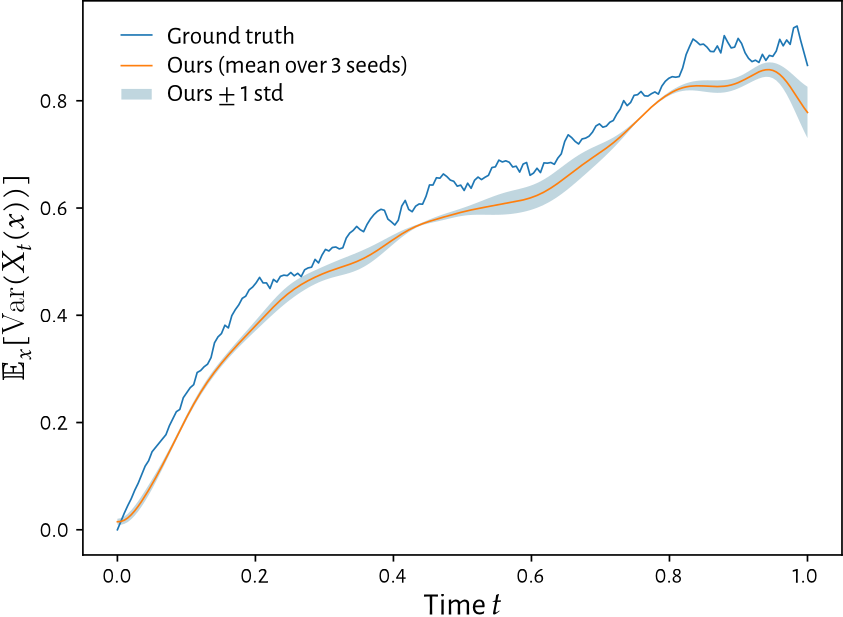}
    \caption{Spatially averaged variance}
    \label{fig:var_time}
  \end{subfigure}
  \hfill
  \begin{subfigure}[t]{0.338\textwidth}
    \centering
    \includegraphics[width=\textwidth]{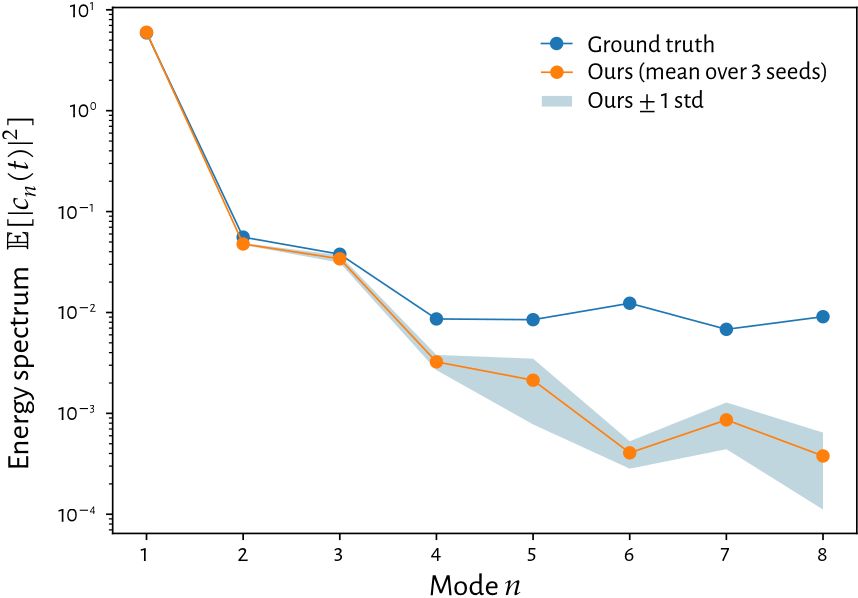}
    \caption{Energy spectrum at final time}
    \label{fig:energy_spec}
  \end{subfigure}
  \hfill
  \begin{subfigure}[t]{0.32\textwidth}
    \centering
    \includegraphics[width=\textwidth]{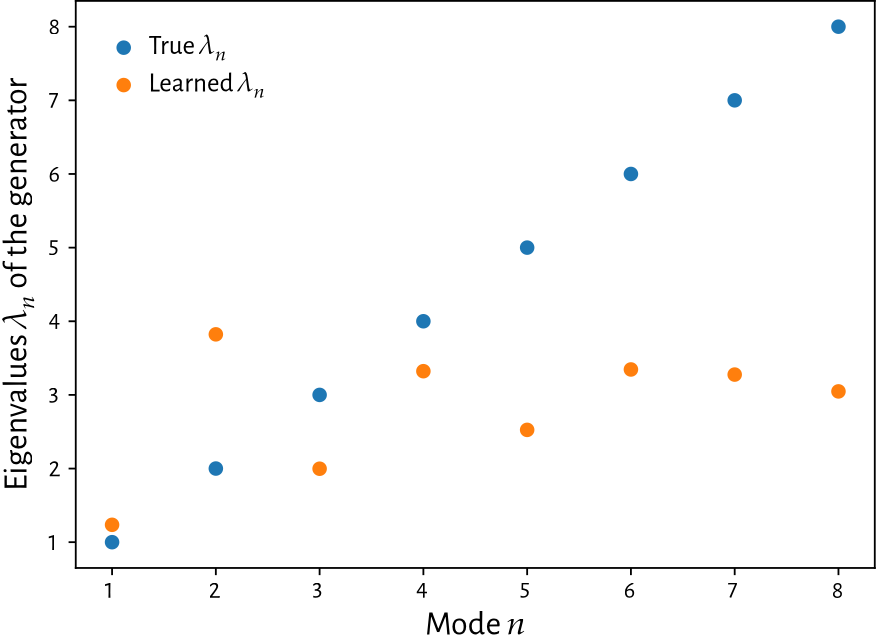}
    \caption{Learned eigenvalues $\lambda_n$}
    \label{fig:lambda}
  \end{subfigure}

  \caption{
  \textbf{Law-level evaluation in Regime A (unbounded domain).}
  \emph{(a)} Temporal evolution of spatially averaged variance
  $\mathbb{E}_x[\mathrm{Var}(X_t(x))]$ (mean $\pm$ std over 3 seeds).
  \emph{(b)} Energy spectrum $\mathbb{E}[|c_n(t)|^2]$ at the final time (mean $\pm$ std over 3 seeds, log scale).
  \emph{(c)} Learned eigenvalues $\lambda_n$ compared to ground truth.
  }
  \label{fig:regimeA_law}
\end{figure*}

\subsection{Regime B: Bounded domain with Dirichlet boundary conditions}

Let \(H := L^2(0,\pi)\) and let \(A\) denote the Laplacian on \((0,\pi)\)
with homogeneous Dirichlet boundary conditions.
The operator \(-A\) admits a complete orthonormal eigenbasis
\((h_n)_{n\ge 1}\) of \(H\) with corresponding eigenvalues
\((\lambda_n)_{n\ge 1}\), where
\[
h_n(x)=\sqrt{\tfrac{2}{\pi}}\sin(nx),
\qquad
-A h_n = \lambda_n h_n, \quad \lambda_n = n^2 .
\]

We study the stochastic heat equation
\begin{equation}
\label{exp:eq:stochheat}
\begin{cases}
dX_t = A X_t\,dt + dW_t, \quad t\in[0,T],\\
X_0 = u_0 \in H,
\end{cases}
\end{equation}
where \((W_t)_{t\ge 0}\) is an \(H\)-valued \(Q\)-Wiener process.

We fix the final time \(T=1\) and choose a deterministic initial condition
\(u_0 \in H\) given by
\[
u_0(x) = x(\pi - x)\exp\!\left(-\frac{(x-0.5\pi)^2}{\sigma^2}\right),
\qquad \sigma=0.5.
\]

The time interval \([0,T]\) is discretized using \(M_2=200\) points, and the spatial
domain \((0,\pi)\) using \(M_3=100\) grid points.
These discretization parameters are chosen according to the parabolic scaling
of the heat operator, such that \((\Delta x)^2 \approx c\,\Delta t\) with \(c=0.2\).
All solution trajectories are generated using the numerical procedure described
in the data generation setup.

Table~\ref{tbl:exp:regime_b:results} reports prediction accuracy for Regime~B.
We evaluate DeepONet, FNO, NSPDE, and DLR-Net using the publicly available SPDEBench baseline library \cite{Li25spdebench_code}, following the experimental setup provided therein.
Minor modifications were required to align data formats; all model architectures and hyperparameters were kept consistent with the provided configurations.
Our method substantially improves both relative $L^2$ error and RMSE under identical observation assumptions.

\begin{table}[H]
  \caption{Prediction accuracy on Regime B (stochastic heat equation) measured by relative $L^2$ error and RMSE (lower is better). Reported values are mean $\pm$ standard deviation over test trajectories, averaged over 3 random model initializations.}
  \label{tbl:exp:regime_b:results}
  \begin{center}
    \begin{small}
      \begin{sc}
        \scalebox{0.90}{
        \begin{tabular}{lr|cc@{}}
          \toprule
          \textbf{Model} & \#Para & rel. $L^2 \;(\times 10^{-1})$ & RMSE $(\times 10^{-1})$ \\
          \midrule
          DeepONet & 11292672 & 4.726 $\pm$ 0.015 & 3.226 $\pm$ 0.021 \\
          FNO      & 4929249  & 4.783 $\pm$ 0.013 & 3.299 $\pm$ 0.014 \\
          NSPDE    & 3283457  & 4.764 $\pm$ 0.005 & 3.275 $\pm$ 0.004 \\
          DLR-Net  & 133178   & 4.748 $\pm$ 0.010 & 3.255 $\pm$ 0.014 \\
          \midrule
          Ours     & 550077   & \textbf{1.114} $\pm$ 0.018 & \textbf{0.679} $\pm$ 0.012 \\
          \bottomrule
        \end{tabular}
        }
      \end{sc}
    \end{small}
  \end{center}
  \vskip -0.2in
\end{table}

To assess approximation of the solution law in Regime~B, we evaluate the same
second-order statistics as in Regime~A.
The temporal evolution of the spatially averaged variance and the induced modal
energy spectrum exhibit close agreement between generated and reference
trajectories, indicating that the learned model captures the correct second-order
stochastic structure in the bounded-domain setting.
Corresponding law-level diagnostics, including a cross-regime comparison of the
learned noise amplitudes, are reported in Appendix~\labelcref{app:sec:addlawresults}, \cref{app:fig:regimeB_law} and \cref{app:fig:q_comparison}.
\vspace*{-.3cm}
\paragraph{Reproducibility.}
An anonymized implementation of the proposed method, together with scripts used in the experiments, is provided in the supplementary material. The full codebase will be released publicly upon acceptance.

%% file: sections/discussion.tex
\paragraph{Conclusion.}
We present a structure-aware latent-variable framework for learning the law of linear SPDEs with additive Gaussian forcing from spatiotemporal solution observations alone. The method combines a spectral Galerkin reduction with a Wiener-Ito chaos representation and exploits first-order chaos closure to obtain deterministic ODE dynamics for the propagator coefficients. This yields an interpretable separation between deterministic evolution and latent stochastic excitation, enabling variational inference of model parameters and realization-specific forcing without observing noise paths. Across bounded and unbounded one-dimensional regimes, the approach improves trajectory accuracy and reproduces law-level second-order statistics, indicating that it captures intrinsic stochastic variability rather than only fitting individual sample paths.
\vspace*{-.3cm}
\paragraph{Outlook.}
The formulation suggests several extensions. First, one can move beyond the linear-additive setting by using higher-order chaos truncations for nonlinear dynamics or multiplicative while preserving the separation between temporal dynamics and stochastic coordinates. Second, replacing the current encoder with architectures that exploit richer spatial structure would enable application to two- and three-dimensional domains with the same latent propagator model and reconstruction operator. Finally, stronger priors or physics-informed constraints on spectral parameters and noise structure may improve data efficiency and robustness under sparse, irregular, or noisy observations.
\vspace*{-.3cm}
\paragraph{Limitations.}
The approach relies on first-order Wiener chaos closure, which holds for linear SPDEs with additive Gaussian forcing; more general SPDE classes are possible but may substantially increase latent dimensionality and training complexity. Moreover, individual spectral parameters may not be fully identifiable from partial observations, since process noise and observation noise can trade off in the likelihood. Our validation is on synthetic data, and real-world deployments will require careful handling of model mismatch, boundary effects, and observation operators. Despite these limitations, the results indicate that combining classical SPDE structure with variational inference provides a principled route toward law-aware data-driven learning of stochastic dynamics.
\vspace*{-.3cm}
\paragraph{Broader Impact.}
This work focuses on methodological advances for learning the law of linear SPDEs from spatiotemporal observations and is evaluated on synthetic data. We do not anticipate direct societal impacts. Potential applications to real-world stochastic systems would require domain-specific validation beyond the scope of this work.

%% file: sections/suppmat.tex
\section{Evaluation Metrics}
\label{sec:evaluation_metrics}

We evaluate predictive performance using second-order statistics that characterize
the distributional structure of the solution process, i.e., we assess whether the learned model reproduces the law-level variability of $X_t$ rather than only trajectory-wise accuracy.

\subsection{Energy spectrum}

Let \( X_t \in H \) admit the truncated spectral representation \footnote{In practice, $X_t$ denotes the projected field onto the first $N$ modes, $X_t^{(N)} := P_N X_t$.}
\[
X_t(x)
=
\sum_{n=1}^N c_n(t)\, h_n(x),
\]
where \( \{h_n\}_{n \geq 1} \) is an orthonormal basis of the Hilbert space \( H \).
The corresponding random coefficients are defined by the spatial inner product
\[
c_n(t)
=
\langle X_t, h_n \rangle_{H}.
\]

The \emph{energy spectrum} at time \(t\) is defined as
\[
E_n(t)
:=
\mathbb{E}\!\left[ |c_n(t)|^2 \right],
\qquad n = 1,\dots,N,
\]
where the expectation is taken with respect to the underlying probability space.
This quantity measures the contribution of the \(n\)-th spatial mode to the total
expected energy of the solution, and satisfies $\sum_{n=1}^N E_n(t) = \mathbb{E}\|X_t\|_H^2$ for the truncated expansion by Parseval's identity.

\subsection{Spatially averaged variance}

For each fixed \( (t,x) \), the pointwise variance of the solution is
\[
\mathrm{Var}(X_t(x))
=
\mathbb{E}\!\left[ |X_t(x)|^2 \right]
-
\bigl|\mathbb{E}[X_t(x)]\bigr|^2.
\]

We evaluate the temporal evolution of the spatially averaged variance
\[
\mathbb{E}_x\!\left[\mathrm{Var}(X_t(x))\right],
\]
where the spatial expectation \( \mathbb{E}_x \) is interpreted according to the
underlying spatial domain:
\[
\mathbb{E}_x[f(x)]
=
\begin{cases}
\displaystyle
\frac{1}{|\mathcal D|}
\int_{\mathcal D} f(x)\,\mathrm{d}x,
& \text{for bounded domains } \mathcal D,
\\[1.2ex]
\displaystyle
\int_{\mathbb R} f(x)\,\gamma(x)\,\mathrm{d}x,
& \text{for weighted spaces } H = L^2(\mathbb R,\gamma),
\end{cases}
\]
with \( \gamma \) a probability density.

If \( \{h_n\}_{n \geq 1} \) is orthonormal in \( H \), this quantity admits the
equivalent spectral representation for the truncated expansion
\[
\mathbb{E}_x\!\left[\mathrm{Var}(X_t(x))\right]
=
\sum_{n=1}^N \mathrm{Var}\!\left( c_n(t) \right).
\]

\subsection{Numerical estimation}

All expectations are approximated using Monte Carlo averages over independent
solution realizations, and spatial integrals are computed via numerical quadrature
on the observation grid, unless stated otherwise, we estimate these statistics from unconditional samples generated by the learned model and compare them to the same estimates computed from reference trajectories.

\section{Training Details and Hyperparameters}
\label{app:sec:hyperparameters}
\begin{table}[H]
\centering
\caption{Optimization and training hyperparameters used for all experiments unless stated otherwise.}
\label{tab:hyperparams}
\begin{small}
\begin{tabular}{ll}
\toprule
\textbf{Category} & \textbf{Setting} \\
\midrule
\multicolumn{2}{l}{\emph{Optimization}} \\
Batch size & 40 \\
Weight decay & $1\times10^{-4}$ \\
Learning rate (encoder) & $1\times10^{-3}$ \\
Learning rate (ODE parameters) & $2\times10^{-2}$ \\
Learning rate (decoder variance) & $1\times10^{-3}$ \\
\midrule
\multicolumn{2}{l}{\emph{Architecture}} \\
Encoder latent dimension & 256 \\
\midrule
\multicolumn{2}{l}{\emph{Regularization}} \\
KL weight (initial latent state) & $7\times10^{-2}$ \\
KL weight (chaos variables) & $1.3$ \\
\bottomrule
\end{tabular}
\end{small}
\end{table}

\section{Computational resources}
All experiments were conducted on an Ubuntu 22.04.5 system with Linux kernel 5.15.0-164-generic, equipped with 32 Intel Core i9-9960X CPU cores, 128\,GB RAM, and four NVIDIA GeForce RTX 2080~Ti GPUs (11\,GB VRAM each).

\section{Additional Law-Level Results}
\label{app:sec:addlawresults}

\begin{figure*}[h]
  \centering

  \begin{subfigure}[t]{0.32\textwidth}
    \centering
    \includegraphics[width=\textwidth]{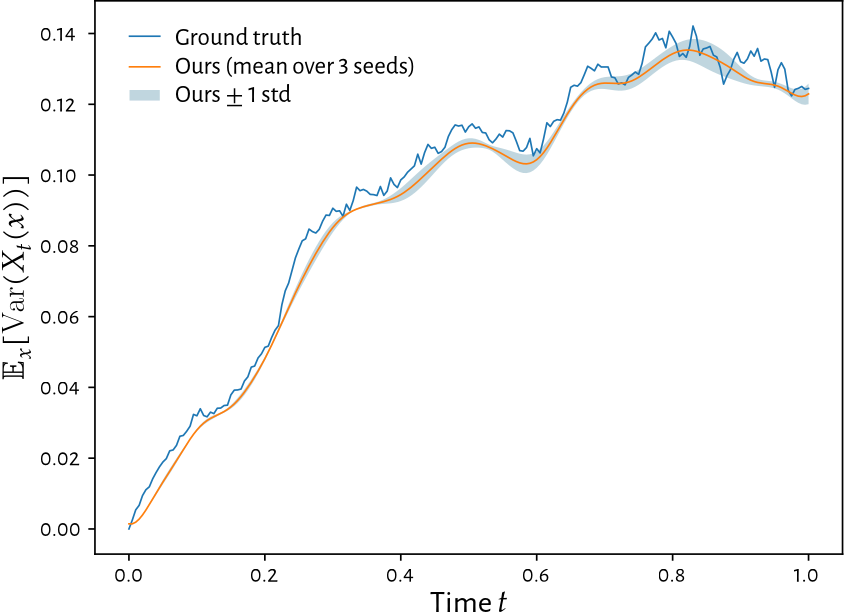}
    \caption{Spatially averaged variance}
    \label{app:fig:var_time}
  \end{subfigure}
  \hfill
  \begin{subfigure}[t]{0.338\textwidth}
    \centering
    \includegraphics[width=\textwidth]{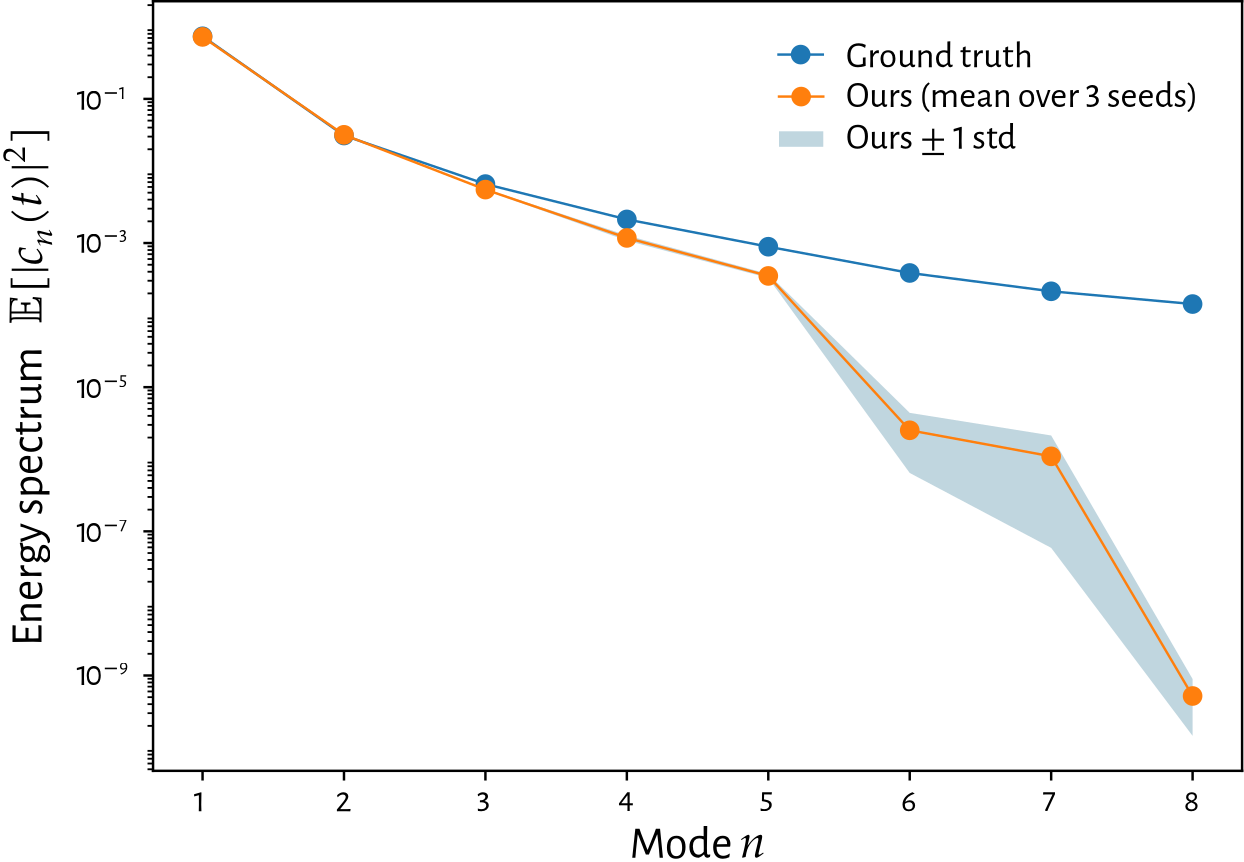}
    \caption{Energy spectrum at final time}
    \label{app:fig:energy_spec}
  \end{subfigure}
  \hfill
  \begin{subfigure}[t]{0.32\textwidth}
    \centering
    \includegraphics[width=\textwidth]{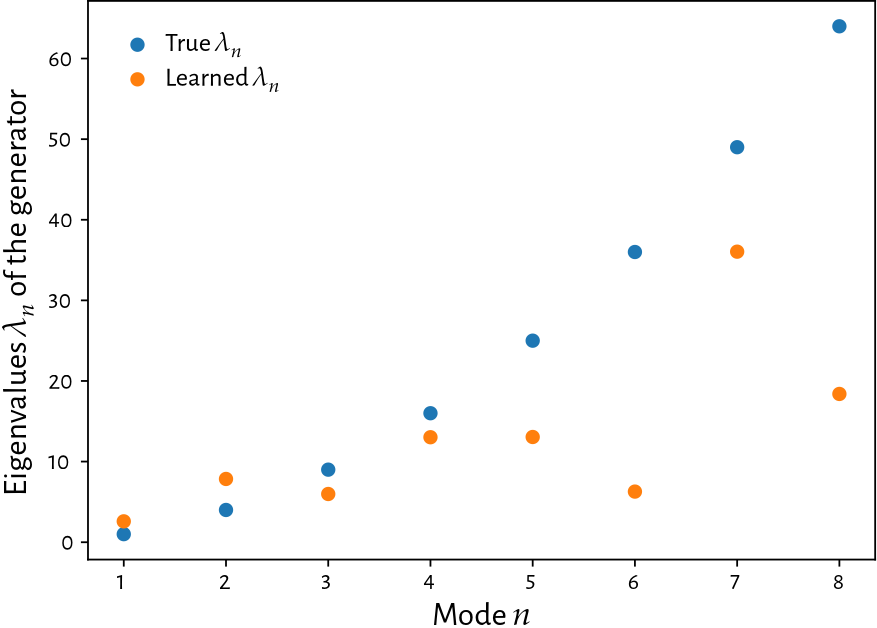}
    \caption{Learned eigenvalues $\lambda_n$}
    \label{app:fig:lambda}
  \end{subfigure}

  \caption{
  \textbf{Law-level evaluation in Regime B (bounded domain).}
  \emph{(a)} Temporal evolution of spatially averaged variance
  $\mathbb{E}_x[\mathrm{Var}(X_t(x))]$ (mean $\pm$ std over 3 seeds).
  \emph{(b)} Energy spectrum $\mathbb{E}[|c_n(t)|^2]$ at the final time (log scale; mean $\pm$ std over 3 seeds).
  \emph{(c)} Learned eigenvalues $\lambda_n$ compared to ground truth.
  }
  \label{app:fig:regimeB_law}
\end{figure*}

\begin{figure}[h]
  \centering
  \begin{subfigure}{0.35\linewidth}
    \centering
    \includegraphics[width=\linewidth]{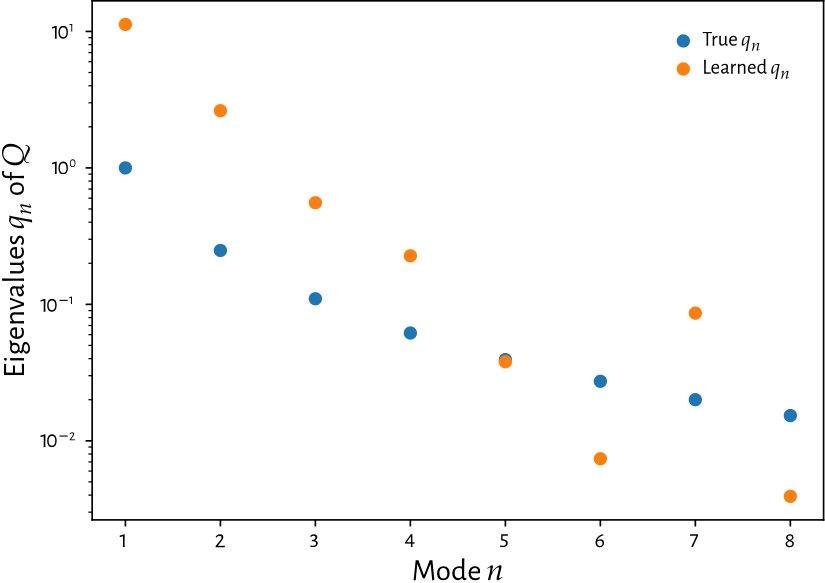}
    \caption{Regime A (unbounded domain)}
  \end{subfigure}
  \hspace{0.02\linewidth}
  \begin{subfigure}{0.35\linewidth}
    \centering
    \includegraphics[width=\linewidth]{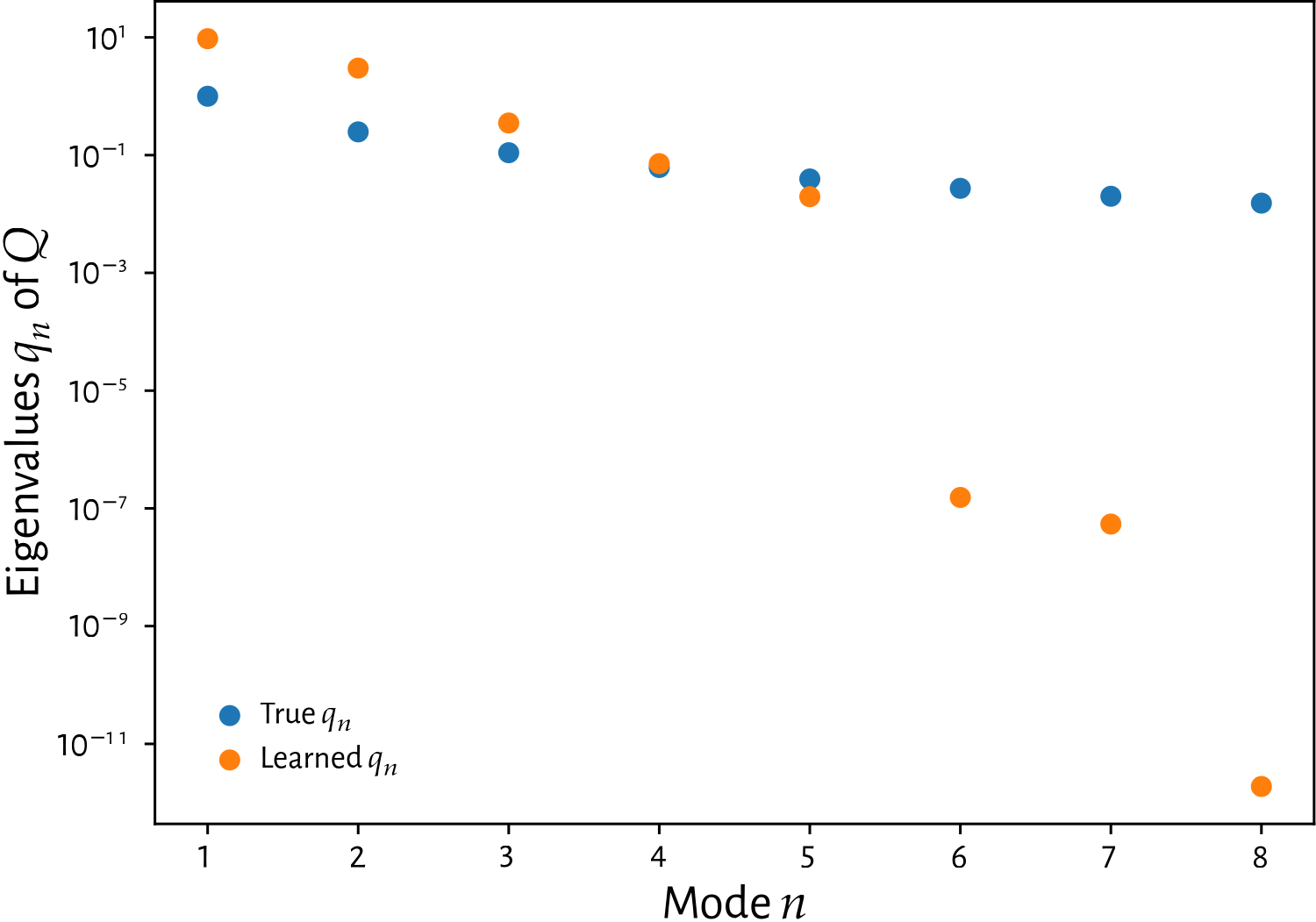}
    \caption{Regime B (bounded domain)}
  \end{subfigure}
  \caption{
  True and learned noise amplitudes \(q_n\) shown on a log--log scale.
  In Regime~A, the learned coefficients recover the correct scale and decay behavior.
  In Regime~B, higher-mode noise amplitudes are suppressed.
  }
  \label{app:fig:q_comparison}
\end{figure}

Figure~\ref{app:fig:q_comparison} compares the learned noise amplitudes across
both regimes. While Regime~A exhibits accurate recovery across all modes, in the
bounded-domain setting of Regime~B the learned noise amplitudes decay more rapidly,
suggesting reduced identifiability of higher modes.